\documentclass[letterpaper, 10 pt, conference]{IEEEtran}
\IEEEoverridecommandlockouts
\usepackage{balance}
\usepackage{cite}
\usepackage{amsmath,amssymb,amsfonts,bbm,mathtools}
\usepackage{hyperref}
\usepackage[ruled, linesnumbered]{algorithm2e}
\usepackage{graphicx,subcaption}
\usepackage{textcomp}
\usepackage{xcolor}
\usepackage{comment}
\usepackage{multirow}
\let\labelindent\relax
\usepackage{enumitem}
\def\BibTeX{{\rm B\kern-.05em{\sc i\kern-.025em b}\kern-.08em
    T\kern-.1667em\lower.7ex\hbox{E}\kern-.125emX}}
\begin{document}

\title{\LARGE \bf Real-Time Constrained 6D Object-Pose Tracking of An In-Hand Suture Needle for Minimally Invasive Robotic Surgery}

\author{
Zih-Yun Chiu$^1$, Florian Richter$^1$ \IEEEmembership{Student Member, IEEE}, and Michael C. Yip$^1$ \IEEEmembership{Senior Member, IEEE}
\thanks{This project was funded by the US Army Telemedicine and Advanced Technologies Research Center and NSF CAREER award \#2045803. F. Richter was supported on an NSF Graduate Research Fellowship.}
\thanks{
$^1$Zih-Yun Chiu, Florian Richter, and Michael Yip are with the Electrical and Computer Engineering Dept., University of California San Diego, La Jolla, CA 92093 USA. {\tt\small \{zchiu, frichter, yip\}@ucsd.edu}}%
}

\maketitle
\begin{abstract}
Autonomous suturing has been a long-sought-after goal for surgical robotics. Outside of staged environments, accurate localization of suture needles is a critical foundation for automating various suture needle manipulation tasks in the real world.
When localizing a needle held by a gripper, previous work usually tracks them separately without considering their relationship. 
Because of the significant errors that can arise in the stereo-triangulation of objects and instruments, their reconstructions may often not be consistent. 
This can lead to unrealistic tool-needle grasp reconstructions that are infeasible.
Instead, an obvious strategy to improve localization would be to leverage constraints that arise from contact, thereby constraining reconstructions of objects and instruments into a jointly feasible space.
In this work, we consider feasible grasping constraints when tracking the 6D pose of an in-hand suture needle. 
We propose a reparameterization trick to define a new state space for describing a needle pose, where grasp constraints can be easily defined and satisfied. 
Our proposed state space and feasible grasping constraints are then incorporated into Bayesian filters for real-time needle localization. 
In the experiments, we show that our constrained methods outperform previous unconstrained/constrained tracking approaches and demonstrate the importance of incorporating feasible grasping constraints into automating suture needle manipulation tasks. 
%Finally, we reconstruct the tracked tool poses for real endoscopic ex-vivo datasets and show that higher accuracy when accounting for feasible grasps than compared to previous unconstrained approach.
\end{abstract}

\section{Introduction}
Automating surgical procedures such as suturing has drawn increased interest within the robotics community during the past two decades~\cite{yip2019robot}. 
The advantage of automation is that it relieves surgeons from time-consuming, tedious, and challenging tasks that often emerge in Minimally Invasive Surgeries~\cite{garcia1998manual,hubens2003performance,corcione2005advantages}. 
One of the key components of achieving autonomous procedures is the accurate localization of surgical instruments in the surgical scene~\cite{li2020super,lu2021super,richter2021robotic}.
This localization ability serves as the foundation for automating various surgical tasks in previous work, including needle regrasping~\cite{chiu2021bimanual,wilcox2022learning}, knot tying~\cite{chow2013improved,lu2018vision}, and blood suction~\cite{richter2021autonomous}.

\begin{figure}[t!]
    \centering
    \vspace{2mm}
    \includegraphics[width=\linewidth]{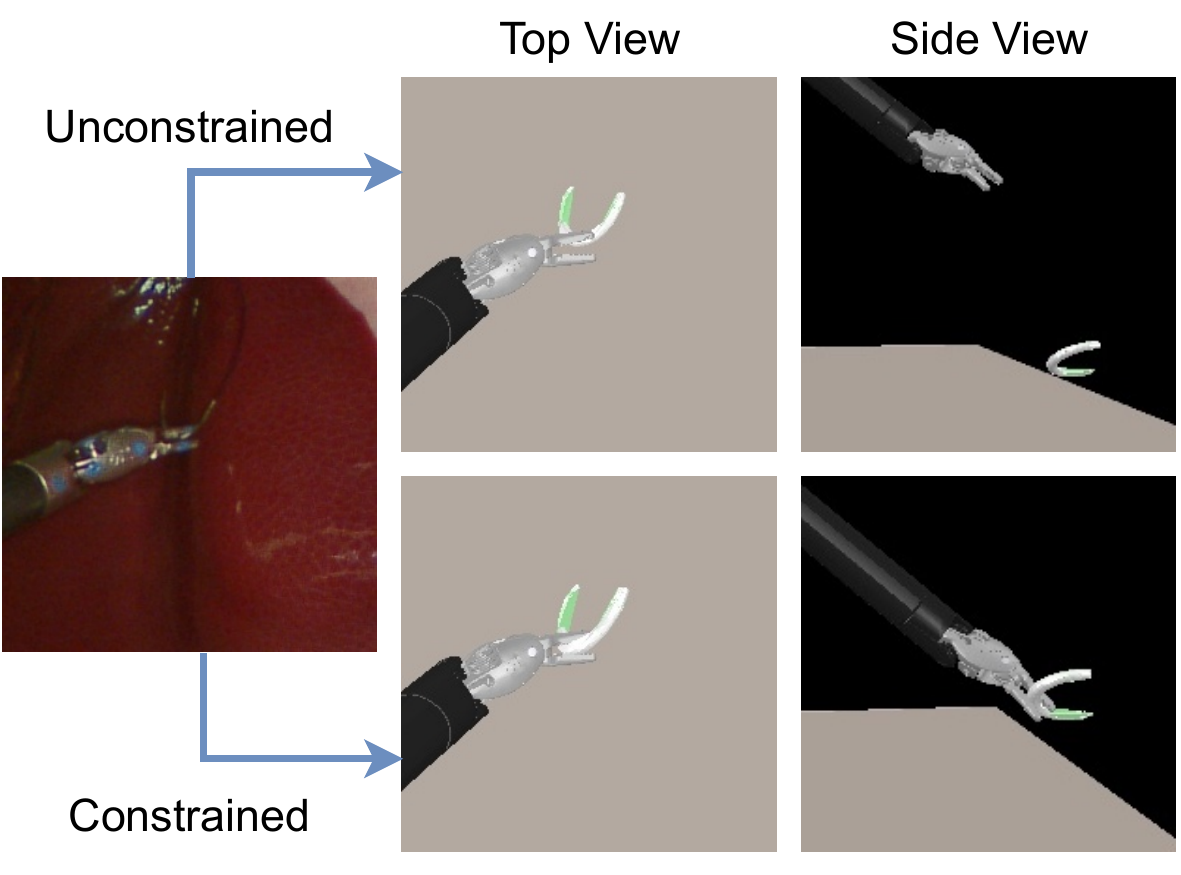}
    \caption{Live image of a daVinci robot instrument grasping a suture needle, top and side views of tool reconstruction from unconstrained and our constrained needle tracking results. The scene reconstruction of our constrained method is always feasible. This feasibility is not ensured by unconstrained approaches, even when the top view of tool reconstruction aligns well with the live image.}
    \label{fig:cover_image}
    \vspace{-3mm}
\end{figure}

A surgical scene often contains multiple surgical instruments, and previous studies localize them separately without considering their physical interactions~\cite{iyer2013single,ferro2017vision,sen2016automating,chiu2021bimanual,wilcox2022learning}. 
This can lead to unrealistic environmental reconstruction when combining tracking results of different tools. 
For example, if a needle is held by a surgical manipulator, tracking them separately can result in the needle being in a non-feasible grasp configuration (e.g., in-collision or floating, as shown in Fig. \ref{fig:cover_image}).
This can be a dangerous situation because dropping needles can result in damage to surrounding tissue and additional trauma with repetitive needle pick-up.
Therefore, in this work, we focus on considering the physical interactions between a suture needle and a surgical manipulator to ensure feasible results in real-time tracking of an in-hand suture needle.
Real-time localization is necessary since, in practice, grasping a needle causes it to re-orient in the gripper, and further re-orientation or slippage can happen once the needle interacts with the environment.

%might add 3D reconstruction to it

\subsection{Related Work}

Current literature on suture needle localization mostly focuses on the features of a needle extracted from camera data and assumes the needle can be anywhere in the space. Several methods reconstruct the pose of a static needle by observing detected markers or learned segmentation~\cite{iyer2013single,d2018automated,wilcox2022learning}. 
Others have considered uncertainty in the features and motions of a needle and use Bayesian filters with different observation models to track its pose~\cite{kurose2013preliminary,ferro2017vision,sen2016automating}. 
However, these methods do not consider the physical interactions between the needle and the environment and thus do not guarantee that the needle pose is feasible.

Some studies in suture needle localization consider the physical interactions between a surgical manipulator and a needle when the manipulator holds the needle. 
One way is to perform tracking and assume that the configuration between the needle and the manipulator tip is known and remains unchanged over time~\cite{ozguner2018three}. 
Then the robot Jacobian and joint-sensor readings are used to estimate the motions of the needle. 
However, getting to this known state is nontrivial, as grasping a needle itself is a non-deterministic action, and grasp pose is situation-dependent, such as during regrasping~\cite{chiu2021bimanual}. 
Thus, the work in \cite{chiu2021markerless} does not assume a known configuration of the needle held by an end-effector, and its motions are estimated by a tool-tracking method~\cite{richter2021robotic} that tracks the pose of the end-effector. 
These approaches take into account that the needle should move concurrently with the gripper when held by it. 
Nonetheless, they do not ensure that the suture needle pose lies inside the feasible grasping manifold of the gripper.%, especially after adding the uncertainty in the motions measured from the joint sensors of the robot. 
% Therefore, the tracked needle poses can still be in a collision or not properly attached to the gripper.
% This can be a dangerous situation as dropping needles can result in damage to surrounding tissue and additional trauma with repetitive needle pick-up.

Tracking the poses of an in-hand needle is a constrained pose tracking problem, where the needle should always lie inside the feasible grasping manifold of the gripper. 
However, there is no unified approach to define a feasible grasping manifold since grippers and grasped objects can be in arbitrary shapes, making this task highly nonlinear.
To incorporate constraints into nonlinear tracking problems, previous work follow two approaches: acceptance/rejection sampling~\cite{lang2007bayesian} and optimization~\cite{kong1994sequential,zhao2014constrained}.
Acceptance/rejection methods are known to reduce the diversity of the tracked pose~\cite{zhao2014constrained} and require an excessive number of feasibility checks, making them not desirable for real-time tracking~\cite{hu2020particle}.
On the other hand, optimization methods project the estimated pose onto a feasible manifold.
However, they require the manifold to be defined as equality or inequality constraints~\cite{zhao2014constrained,hu2020particle}, and describing the feasible grasping manifold in such a way would be highly nontrivial.

\subsection{Contributions}

In this work, we achieve state-of-the-art performance for real-time suture needle tracking in robotic surgery by incorporating grasping constraints.
To this end, we present the following novel contributions:
\begin{enumerate}
    \item the first approach to probabilistically track a suture needle in real-time with grasping constraints,
    \item a state-space to describe a grasped suture needle for efficient sampling on the feasible grasping manifold,
    \item and a comparison of Bayesian filter approaches that incorporate the grasping constraints.
\end{enumerate}

The proposed methods are evaluated in both simulation and real-world environments.
In simulation environments, we demonstrate that our proposed methods outperform other unconstrained/constrained tracking approaches. 
Moreover, we evaluate different tracking methods on the suture needle regrasping task~\cite{chiu2021bimanual,wilcox2022learning}. 
The results indicate that incorporating grasping constraints makes the regrasping policy more robust to noise in detections. 
In real-world environments, we use marker-less feature detections from a Deep Neural Network (DNN) as needle observations and reconstruct the tracked tool poses from ex-vivo images. 
An example is shown in Fig. \ref{fig:cover_image}. 
The results demonstrate that our constrained approach ensures a feasible estimated pose, and an unconstrained method can lead to unrealistic reconstructions. 

\section{Methods}

\subsection{Problem Formulation}

We aim to solve the in-hand suture needle pose, $\mathbf{s}_t$, tracking problem probabilistically from a sequence of observations, $\mathbf{o}_{0:t}$, which can be formulated as:
\begin{equation}
\begin{aligned}
    \textrm{Track} \quad & p_{t|t}(\mathbf{s}_t) \coloneqq p(\mathbf{s}_t | \mathbf{a}_{0:t-1}, \mathbf{o}_{0:t}) \\
    \textrm{s.t.} \quad & \mathbf{s}_t \in \mathcal{F}_t \\
    \textrm{where} \quad &
    \mathbf{s}_t = f(\mathbf{s}_{t-1}, \mathbf{a}_{t-1}, \mathbf{w}_{t-1}) \sim p_f(\cdot | \mathbf{s}_{t-1}, \mathbf{a}_{t-1}) \\
    & \mathbf{o}_t = h(\mathbf{s}_t, \mathbf{v}_t) \sim p_h(\cdot | \mathbf{s}_t)
    \label{equ:cBF_formulation}
\end{aligned}
\end{equation}
where $\mathcal{F}_t$ is the feasible grasping space, $f(\cdot)$ and $h(\cdot)$ are the motion and observation models with noise $\mathbf{w}_{t-1}$ and $\mathbf{v}_t$ respectively, and $\mathbf{a}_{t-1}$ is the action applied to the suture needle.
% \begin{equation}
% \begin{aligned}
%     \textrm{Track} \quad & p_{t|t-1}(\mathbf{s}_t) \text{ and } p_{t|t}(\mathbf{s}_t), t \in \{1, \dots, T\} \subseteq \mathbb{N} \\
%     \textrm{s.t.} \quad & \mathbf{s}_t \in \mathcal{F}_t, \forall t \\
%     \textrm{where} \quad & p_{t|t-1}(\mathbf{s}_t) \coloneqq p(\mathbf{s}_t | \mathbf{a}_{0:t-1}, \mathbf{o}_{0:t-1}) \\
%     & p_{t|t} \coloneqq p(\mathbf{s}_t | \mathbf{a}_{0:t-1}, \mathbf{o}_{0:t}) \\
%     & \mathbf{s}_t = f(\mathbf{s}_{t-1}, \mathbf{a}_{t-1}, \mathbf{w}_{t-1}) \sim p_f(\cdot | \mathbf{s}_{t-1}, \mathbf{a}_{t-1}) \\
%     & \mathbf{o}_t = h(\mathbf{s}_t, \mathbf{v}_t) \sim p_h(\cdot | \mathbf{s}_t)
%     \label{equ:cBF_formulation}
% \end{aligned}
% \end{equation}
% The goal is to keep tracking the posterior probability of the needle state at every time step, $\mathbf{s}_t$, given the previous actions and observations, and $\mathbf{s}_t$ should always belong to the space $\mathcal{F}_t$. 
% In this mathematical formulation, $\mathbf{a}_t, \mathbf{o}_t, \mathbf{w}_t$, and $\mathbf{v}_t$ are the action, observation, motion noise, and observation noise at the time step $t$ respectively. 
% The functions $f$ and $h$ are the possibly nonlinear motion and observation models, and $p_f$ and $p_h$ are their equivalent probability density functions respectively.

In our task, $\mathcal{F}_t$ in (\ref{equ:cBF_formulation}) is the feasible grasping manifold of the surgical manipulator that is holding a suture needle at time step $t$. 
Usually, a grasping manifold should consider two feasibility constraints: geometric and dynamic constraints. 
Geometric constraints include~\cite{hasson2019learning}: 
\begin{enumerate}
    \item The object’s surface should be in contact with the gripper’s surface, i.e., $Surface(\mathbf{{s}_t}) \cap Surface(\mathbf{e}_t) \neq \emptyset$, where $\mathbf{e}_t$ is the state of the gripper at time step $t$.
    \item The object should not penetrate with the gripper, i.e., $Interior(\mathbf{s}_t) \cap Interior(\mathbf{e}_t) = \emptyset$. 
\end{enumerate}
Dynamic constraints include that if there is no external force except gravity acting on both the object and the gripper, the linear and angular velocities of the object relative to the gripper should be $\mathbf{0}$. 
Hence, the feasible grasping manifold $\mathcal{F}_t$ can be represented as 
\begin{alignat}{3}
    \quad & \mathcal{F}_t = \{ \mathbf{s}_t | && \mathbf{s}_t \in \mathcal{G}_t \cap \mathcal{D}_t\}, \label{equ:F_definition}\\
    \textrm{where} \quad & \mathcal{G}_t = \{ \mathbf{s}_t | && Surface(\mathbf{{s}_t}) \cap Surface(\mathbf{e}_t) \neq \emptyset \text{ and } \notag\\
    \quad & && Interior(\mathbf{s}_t) \cap Interior(\mathbf{e}_t) = \emptyset\}, \label{equ:G_definition}\\
    \quad & \mathcal{D}_t = \{ \mathbf{s}_t | && \text{If } ExternalForce \setminus Gravity = \emptyset, \notag\\
    \quad & && LinearVelocity(\mathbf{s}_t, \mathbf{e}_t) = \mathbf{0} \text{ and } \notag\\
    \quad & && AngularVelocity(\mathbf{s}_t, \mathbf{e}_t) = \mathbf{0} \}. \label{equ:D_definition}
\end{alignat}
Due to the special design and property of surgical manipulators and suture needles, we can simplify the requirements of defining the feasible grasping manifold for an in-hand needle. 
More specifically, the dynamic constraints in (\ref{equ:D_definition}) are ignored because (1) a suture needle is very light compared to the gripper, and (2) grippers for surgical manipulators are designed to increase the friction between themselves and the objects they are holding (e.g., Needle Drivers). 
Hence, $\mathcal{F}_t = \mathcal{G}_t, \forall t \in [1, \dots, T]$. 

Since the robot end-effector or the grasped object can have a complex shape, the feasible grasping manifold $\mathcal{F}_t$ in (\ref{equ:F_definition}) is difficult to define for the object pose, $[\mathbf{b}_t^\top \  \mathbf{q}_t^\top]^\top$, where $\mathbf{b}_t \in \mathbb{R}^3$ is the position, and $\mathbf{q}_t \in \mathbb{R}^3$ is the axis-angle orientation. 
The object pose, which is described in a global frame such as the camera frame or in the ego-centric end-effector frame, is often directly used as the state in (\ref{equ:cBF_formulation})~\cite{kurose2013preliminary,sen2016automating,ferro2017vision,ozguner2018three,chiu2021markerless}.
However, without a feasibility-checking library or a physical simulator, it is difficult to tell whether a pose state $\mathbf{s}^p_t = [\mathbf{b}_t^\top \  \mathbf{q}_t^\top]^\top$ belongs to $\mathcal{F}_t$. 
These libraries and simulators require the mesh files of the end-effectors and objects, and multiple proximity queries on geometric models can slow down tracking. 
Moreover, $\mathcal{F}_t$ for the pose state is generally irregular, so randomly sampling a pose state from $\mathcal{F}_t$ can take even more time.

\subsection{Needle Pose Reparameterization}

Instead of defining the state as the needle pose in the camera frame or in the end-effector frame, we propose a \textit{\textbf{reparameterization trick}} that defines a new set of parameters, $(\alpha, w, u, v)$, to describe the pose of an in-hand suture needle in the end-effector frame. 
First, we introduce their \textit{intermediate} parameters, $(\alpha, d, \theta, \phi)$, for the pose which was originally defined in our previous work~\cite{chiu2021bimanual}. 
This set of parameters provides a more intuitive understanding of how to describe the pose of a needle held by a gripper. 
The first parameter, $\alpha \in [\frac{1}{2}\pi, \frac{3}{2}\pi] \subseteq \mathbb{R}$, indicates which point on the needle is grasped, and the other three parameters, $d \in [d_{min}, d_{max}] \subseteq \mathbb{R}$, $\theta \in [\theta_{min}, \theta_{max}] \subseteq \mathbb{R}$, and $\phi \in [\phi_{min}, \phi_{max}] \subseteq \mathbb{R}$, describe the position of the end-effector relative to the grasped point of the needle in the spherical coordinate system. 
% Figure \ref{TODO} illustrates the $(\alpha, d, \theta, \phi)$-parameter for a given pose of an in-hand suture needle. 
% These parameters are used in our previous work~\cite{TODO} to sample an initial configuration of a surgical manipulator holding a needle. 
% Nonetheless, they do not represent the state of a needle in that work.

Although the $(\alpha, d, \theta\, \phi)$-parameter is intuitive, independently sampling for $d, \theta$, and $\phi$ can lead to high biases when transforming from spherical coordinates to Cartesian space~\cite{weisstein2002sphere}, which is where the suture needle pose is defined. 
Thus, the following change of variables is applied~\cite{cundy1989sphere,harman2010decompositional}:
\begin{equation}
    w = d^3, \quad u = \frac{\theta}{2\pi}, \quad v = \frac{1}{2}(\cos\phi + 1), 
    \label{equ:wuv}
\end{equation}
where $w \in [d_{min}^3, d_{max}^3] \subseteq \mathbb{R}$, $u \in [\frac{1}{2\pi}\theta_{min}, \frac{1}{2\pi}\theta_{max}] \subseteq \mathbb{R}$, and $v \in [\frac{1}{2}(\cos\phi_{max}+1), \frac{1}{2}(\cos\phi_{min}+1)] \subseteq \mathbb{R}$.
%\textcolor{red}{See Fig. ? to visualize how the new space, $w, u, v$, is less biased than the original $ d, \theta\, \phi$.}
Then a reparameterized state $\mathbf{s}^r_t$ can be represented as $\mathbf{s}^r_t = [\alpha_t\ w_t\ u_t\ v_t]^\top \in \mathbb{R}^4$.

\begin{figure}[t!]
    \vspace{1.5mm}
    \centering
    \includegraphics[width=0.45\textwidth]{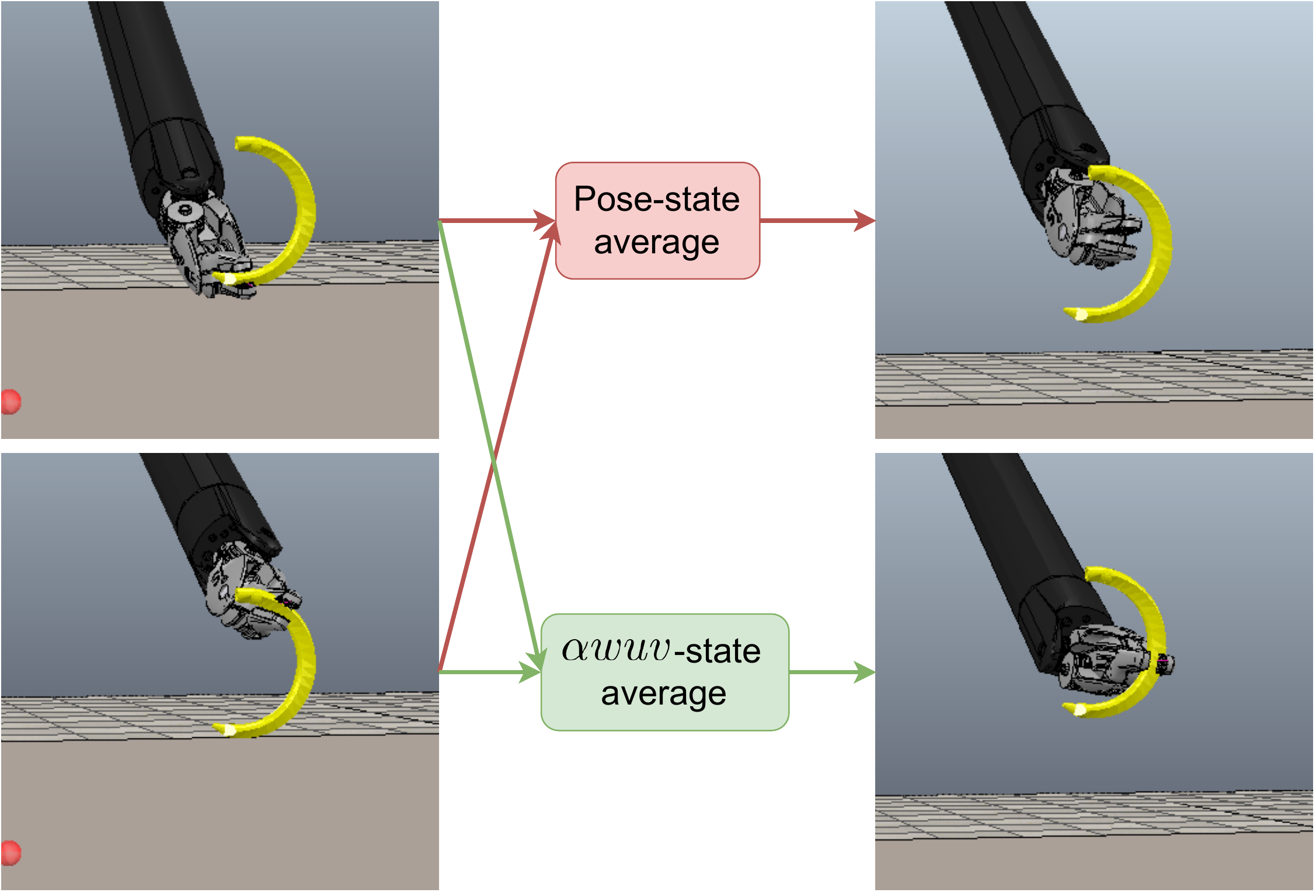}
    \caption{Average of two pose-states and two $(\alpha, w,u,v)$-states. A direct average of two pose-states can lead to an infeasible needle state. However, the direct weighted sum of two $(\alpha, w,u,v)$-states is still feasible since the feasible grasping manifold for $(\alpha, w,u,v)$-states is convex.}
    \label{fig:state_average_example}
    \vspace{-3.5mm}
\end{figure}

Reparameterizing the state of a suture needle as $(\alpha, w, u, v)$ has several benefits. 
First, the geometrically feasible space, $\mathcal{G}_t$ in (\ref{equ:G_definition}), has a concrete definition: 
\begin{align}
    \mathcal{G}_t = \mathcal{G} = \left\{ \right. & 
        \mathbf{s}^r = [\alpha\ \ w\ \ u\ \ v]^\top | \notag\\
        & \alpha \in \left[\frac{\pi}{2}, \frac{3\pi}{2}\right], 
        w \in \left[d_{min}^3, d_{max}^3\right], \notag\\
        & u \in \left[\frac{\theta_{min}}{2\pi}, \frac{\theta_{max}}{2\pi}\right], \notag\\
        & v \in \left[\frac{1}{2}(\cos\phi_{max}+1), \frac{1}{2}(\cos\phi_{min}+1)\right]
    \left. \right\}. 
\end{align}
As long as a state $\mathbf{s}^r$ belongs to $\mathcal{G}$, the needle is in contact with the gripper's surface and not in collision with the end-effector. 
Hence, $\mathbf{s}^r$ is a geometrically feasible state. 
Second, the motion model, $f(\cdot)$ in (\ref{equ:cBF_formulation}), for a $(\alpha, w, u, v)$-state becomes very simple: 
\begin{align}
    \label{equ:s_r_motion_model}
    \mathbf{s}^r_{t+1} & = f(\mathbf{s}^r_t, \mathbf{a}^r_t, \mathbf{w}^r_t) = 
    \textrm{clip}(\mathbf{s}^r_t + \mathbf{a}^r_t + \mathbf{w}^r_t, \mathbf{s}^r_{min}, \mathbf{s}^r_{max}), \\
    \mathbf{s}^r_{min} & = \left[\frac{\pi}{2}\ \ d_{min}^3\ \ \frac{\theta_{min}}{2\pi}\ \ \frac{1}{2}(\cos\phi_{max}+1)\right]^\top, \\
    \mathbf{s}^r_{max} & = \left[\frac{\pi}{2}\ \ d_{max}^3\ \ \frac{\theta_{max}}{2\pi}\ \ \frac{1}{2}(\cos\phi_{min}+1)\right]^\top, 
    \label{equ:s_r_motion_model_end}
\end{align}
where $\mathbf{a}^r_t = [a_{t,\alpha}\ a_{t,w}\ a_{t,u}\ a_{t,v}]^\top \in \mathbb{R}^4$ is the variation applied to $\mathbf{s}^r_t$, and $\mathbf{w}^r_t \in \mathbb{R}^4$ is the motion noise. 
This simple form ensures feasible outputs from the motion model while requiring no time-consuming post-processing, such as rejection sampling or optimization. 
Finally, the geometrically feasible space, $\mathcal{G}$, when using $(\alpha, w, u, v)$-state, is the shape of a hyperrectangle, so it is a convex space.
The convexity of $\mathcal{G}$ enables the direct weighted sum of multiple tracked state candidates to also be feasible, which is important when computing averages.
Meanwhile, for pose-states, the feasible space is usually non-convex due to complex gripper and object shapes in the 3D space, and the weighted average of multiple poses can be infeasible.
Fig. \ref{fig:state_average_example} shows an example of averaging two pose-states and two $(\alpha, w, u, v)$-states. 
The former results in a floating needle, whereas the latter remains feasible grasping. 
% \textcolor{red}{Add a figure showing average in pose space and average in your new space. The average in pose space should be not feasible showing the obvious advantage of your new space.}

\begin{figure}[t!]
    \vspace{1.5mm}
    \centering
    \includegraphics[width=0.35\textwidth]{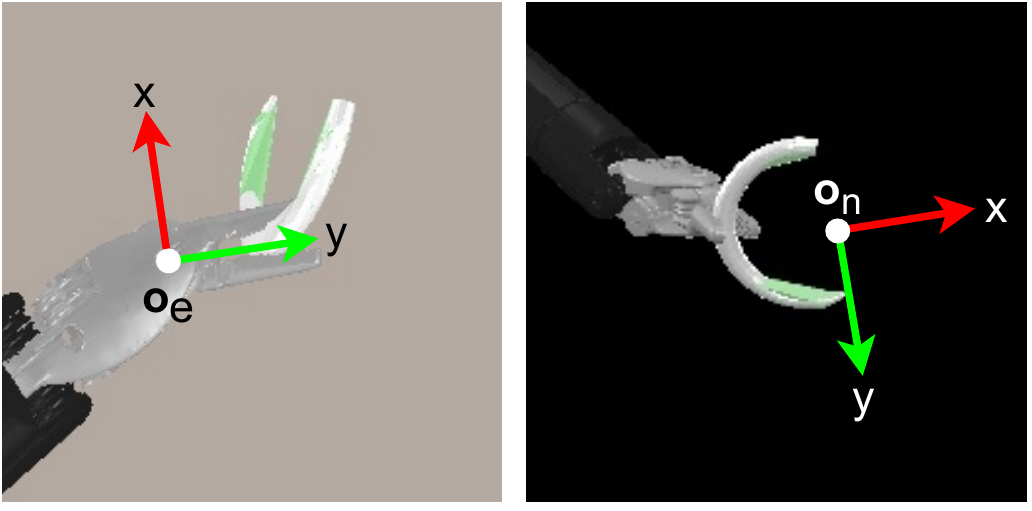}
    \caption{The top and side views of a surgical manipulator stably grasping a suture needle and the coordinate frames used in this work. It can be observed that when a needle is stably grasped, the y-axis of the end-effector frame will pass through at least one point on the needle.}
    \label{fig:stable_grasping_example}
    \vspace{-3.5mm}
\end{figure}

Mappings between a $(\alpha, w, u, v)$-parameter and its corresponding needle pose in the camera frame, $\mathbf{p}^c_n = [\mathbf{b}^{c\top}_n\ \mathbf{q}^{c\top}_n]^\top$, are necessary since the observations of the needle are from the images taken by the camera. 
The steps to transform from $(\alpha, w, u, v)$ to $\mathbf{p}^c_n$ are summarized as follows: 
\begin{enumerate}
    \item Transform $(w, u, v)$ to $(d, \theta, \phi)$ using the inverse of (\ref{equ:wuv}).
    \item Transform $(\alpha, d, \theta, \phi)$ to $\mathbf{p}^n_e$ and $\mathbf{p}^c_n$ using the methods described in~\cite{chiu2021bimanual}, where $\mathbf{p}^n_e = [ \mathbf{b}^{n\top}_e\ \mathbf{q}^{n\top}_e ]^\top$ is the pose of the end-effector in the needle frame. 
\end{enumerate}
Transforming from $\mathbf{p}^c_n$ to $(\alpha, w, u, v)$ requires some properties of an in-hand suture needle.
First, we list the steps of this transformation as follows:
\begin{enumerate}
    \item Calculate $\mathbf{p}^n_e$ using $\mathbf{p}^c_n$ and $\mathbf{p}^c_e$, where $\mathbf{p}^c_e = [ \mathbf{b}^{c\top}_e\ \mathbf{q}^{c\top}_e ]^\top$ is the pose of the end-effector in the camera frame. 
    \item Find out the needle's grasped point in the needle frame, $\mathbf{p}^n_g = [\mathbf{b}^{n\top}_g\ \mathbf{q}^{n\top}_g]^\top$. 
    \item Calculate $(\alpha, d, \theta, \phi)$ using $\mathbf{p}^n_g$ and $\mathbf{p}^n_e$. 
    \item Transform $(d, \theta, \phi)$ to $(w, u, v)$ using (\ref{equ:wuv}).
\end{enumerate}
The first step can be calculated by 
\begin{equation}
    \mathbf{H}(\mathbf{b}^n_e, \mathbf{q}^n_e) = \left( \mathbf{H}(\mathbf{b}^c_n, \mathbf{q}^c_n) \right)^{-1} 
    \mathbf{H}(\mathbf{b}^c_e, \mathbf{q}^c_e). 
\end{equation}
where $\mathbf{H}(\cdot) \in SE(3)$ is the homogeneous transform from position and orientation vectors. 
To obtain $\mathbf{p}^n_g$ in the second step, we use the following property of an in-hand suture needle:
\textit{If a needle is stably grasped by a surgical manipulator, the y-axis of the end-effector frame will pass through the needle’s grasped point.}
An example can be observed in Figure \ref{fig:stable_grasping_example}. 
This property allows us to calculate $\mathbf{p}^n_g$ by 
\begin{align}
    & \mathbf{b}^n_g = \mathbf{b}^n_e + \beta \mathbf{R}^n_{e,y}, \label{equ:b_n_g}\\
    & \mathbf{q}^n_g = \mathbf{0}, \\
    \textrm{where} \quad & \beta = -\left(
        [0\ 0\ 1] \mathbf{b}^n_e
    \right) / 
    \left( [0\ 0\ 1] \mathbf{R}^n_{e,y} \right).
\end{align}
The coefficient $\beta \in \mathbb{R}$ in (\ref{equ:b_n_g}) is obtained by \textit{finding the intersection between $\mathbf{R}^n_{e,y}$ and the xy-plane of the needle frame}, where $\mathbf{R}^n_{e,y} \in \mathbb{R}^3$ is the y-axis of the end-effector frame in the needle frame. 
With $\mathbf{p}^n_g$, $\alpha$ can be calculated by 
\begin{equation}
\vspace{-2mm}
    \alpha = \tan^{-1}\left( \frac{b^n_{g,y}}{b^n_{g,x}} \right), 
    \vspace{1mm}
\end{equation}
where $b^n_{g,x} \in \mathbb{R}$ and $b^n_{g,y} \in \mathbb{R}$ are the x- and y-coordinates of $\mathbf{b}^n_g$~\cite{chiu2021bimanual}. 
To obtain $(d, \theta, \phi)$ in the third step, we first calculate 
\begin{equation}
    \mathbf{b}^g_e = [b^g_{e,x}\ \ b^g_{e,y}\ \ b^g_{e,z}]^\top = \mathbf{b}^n_e - \mathbf{b}^n_g. 
\end{equation}
Then the $(d, \theta, \phi)$ -parameters become~\cite{chiu2021bimanual} 
\begin{equation}
    d = \left\lVert \mathbf{b}^g_e \right\rVert_2, 
    \theta = \tan^{-1}\left( \frac{b^g_{e,y}}{b^g_{e,x}} \right), 
    \phi = \cos^{-1} \left( \frac{b^g_{e,z}}{d} \right). 
\end{equation}
Finally, $(w, u, v)$ in the fourth step can be calculated by (\ref{equ:wuv}). 

\begin{figure*}[t!]
    \vspace{1.5mm}
    \centering
    \begin{subfigure}{0.48\linewidth}
        \centering
        \includegraphics[width=\textwidth]{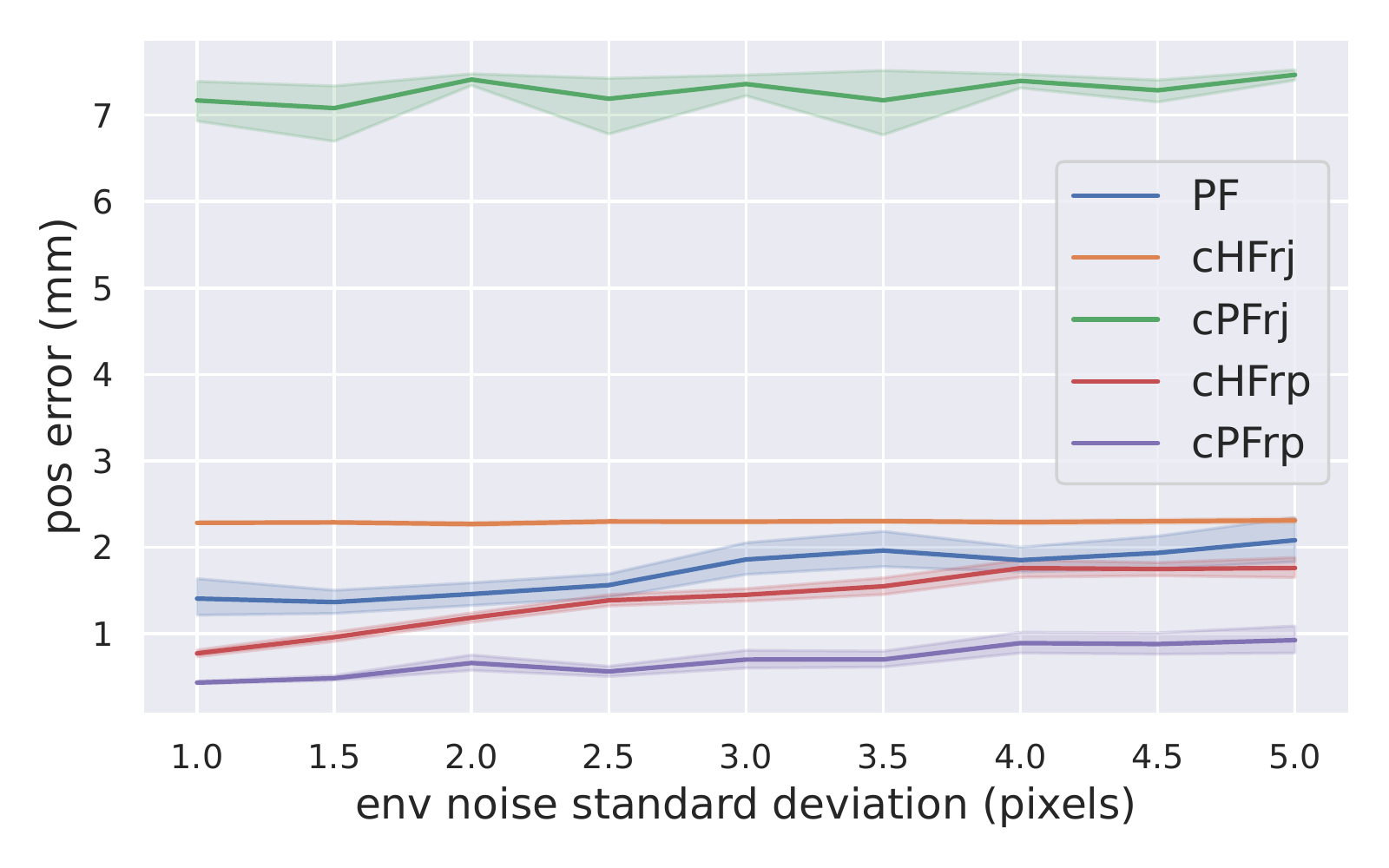}
    \end{subfigure}%
    \begin{subfigure}{0.48\linewidth}
        \centering
        \includegraphics[width=\textwidth]{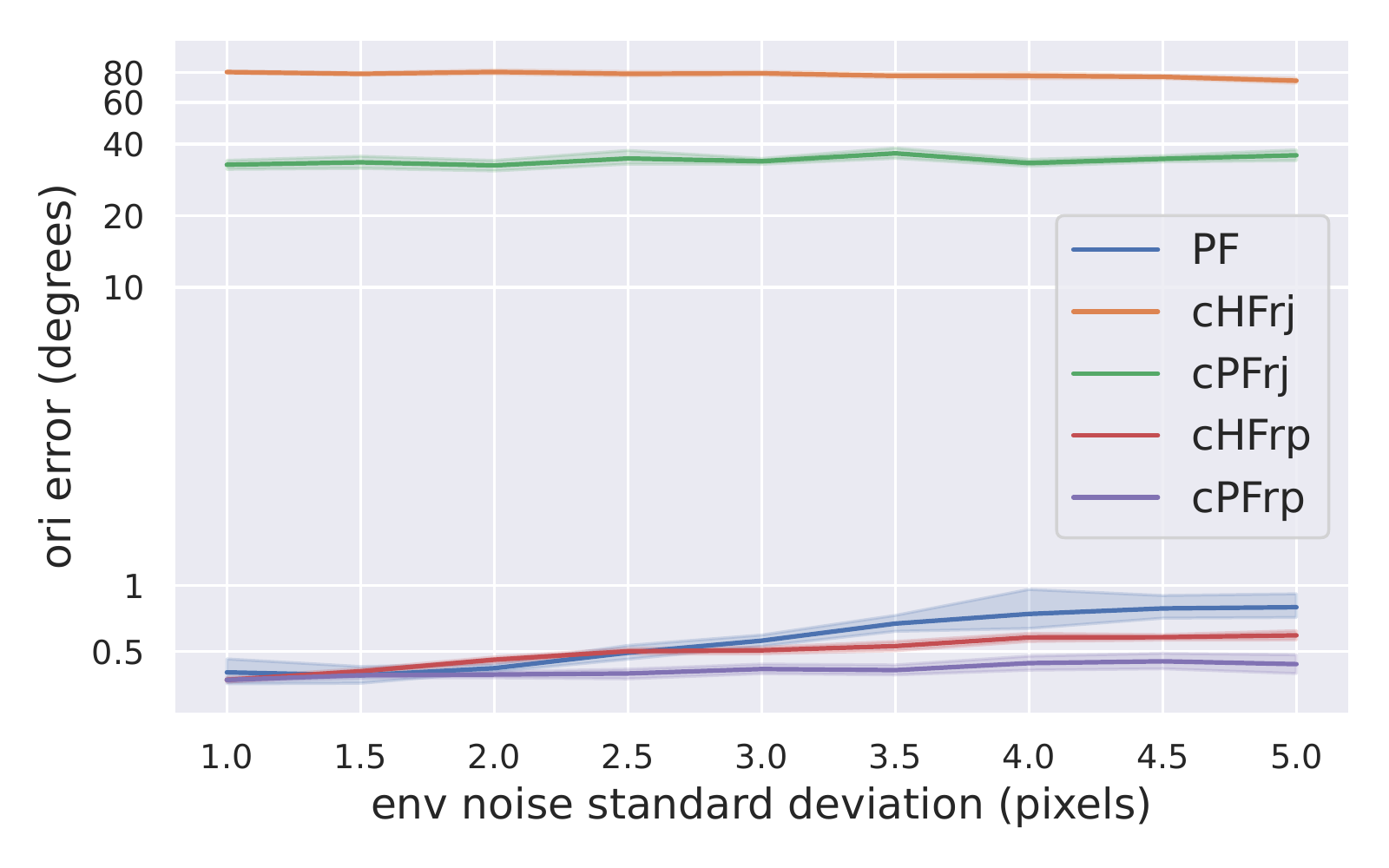}
    \end{subfigure}
    \caption{The positional errors (left) and orientational errors (right, log-scale) of tracking an in-hand suture needle grasped by a moving gripper. Our proposed state-reparameterization methods outperform other approaches since the new state space allows a concrete definition of the feasible grasping manifold and can easily incorporate grasping constraints into tracking.}
    \label{fig:simulation_tracking_errors}
    \vspace{-5.5mm}
\end{figure*}

\subsection{Bayesian Filter with Feasibility Constraints}

\begin{algorithm}[t!]
  \vspace{0.5mm}
  \footnotesize
  \SetAlgoLined
  \KwIn{maximal and minimal values of $(\alpha,w,u,v)$-parameters $\mathbf{s}^r_{min}$ and $\mathbf{s}^r_{max}$, motion noise covariance $\Sigma^r_m$, image observations $\mathbb{I}_{1:T}$, observation noise variance $\sigma^2_o$, needle radius $r$, tracking algorithm HF or PF}
  \KwOut{tracked needle state $\mathbf{s}^r_{1:T}$}
  % \tcp{Initialize States/Particles and Their Weights, (\ref{equ:s_r_definition}) to (\ref{equ:sample_s_r})}
  \tcp{Initialization}
  $\{ \mathbf{s}^r_{0,i} \}_{i=1}^n \leftarrow sampleFeasible(\mathbf{s}^r_{min}, \mathbf{s}^r_{max}$) \\
  \label{alg:initialize_sample}
  $\{ \gamma_{0,i} \}_{i=1}^n \leftarrow \{ \frac{1}{n}, \dots, \frac{1}{n} \}$ \\
  \label{alg:initialize_weights}
  \For{timestep $t = 1,\dots, T$}{
    \tcp{Predict Step}
    \uIf{HF}{
        \label{alg:start_mm}
      $\{ \gamma_{t,i} \}_{i=1}^n \leftarrow Predict(\{ \mathbf{s}^r_{t-1,i}, \gamma_{t-1,i} \}_{i=1}^n, \Sigma^r_m)$ \\
      $\{ \mathbf{s}^r_{t,i} \}_{i=1}^n \leftarrow \{ \mathbf{s}^r_{t-1,i} \}_{i=1}^n$ \\
    }
    \ElseIf{PF}{
      $\{ \mathbf{s}^r_{t,i} \}_{i=1}^n \leftarrow Predict(\{ \mathbf{s}^r_{t-1,i} \}_{i=1}^n, \Sigma^r_m)$ \\
      $\{ \gamma_{t,i} \}_{i=1}^n \leftarrow \{ \gamma_{t-1,i} \}_{i=1}^n$ \\
    }
     \label{alg:end_mm}
     % \tcp{Detect Needle Features}
     \tcp{Update Step}
    $\mathbf{o}_t \leftarrow getDetections(\mathbb{I}_t)$\\
    \label{alg:start_om}
    % \tcp{Transform $(\alpha,w,u,v)$-state to pose-state, (\ref{equ:wuv}) and (\ref{TODO}) to (\ref{TODO})}
    $\{ \mathbf{s}^p_{t,i} \}_{i=1}^n \leftarrow \{ \alpha wuv2pose(\mathbf{s}^r_{t,i}, \mathbf{p}^c_{e,t}, r) \}_{i=1}^n$ \\
    % \tcp{Update}
    $\{ \gamma_{t,i} \}_{i=1}^n \leftarrow weightUpdate(\{ \mathbf{s}^p_{t,i}, \gamma_{t,i} \}_{i=1}^n, \mathbf{o}_t, \sigma^2_o)$ \\
    \label{alg:end_om}
    \If{PF}{
      \label{alg:start_pf_om}
      % \tcp{Reparameterize pose-state, (\ref{equ:wuv}) and (\ref{TODO}) to (\ref{TODO})}
      $\{ \mathbf{s}^r_{t,i} \}_{i=1}^n \leftarrow \{ pose2\alpha wuv(\mathbf{s}^p_{t,i}, \mathbf{p}^c_{e,t}) \}_{i=1}^n$ \\
      \tcp{Stratify Resampling \cite{kitagawa1996monte}}
      \If{$effectiveParticles(\{ \gamma_{t,i} \}_{i=1}^n) < N_{eff}$}{
        $\{ \mathbf{s}^r_{t,i}, \gamma_{t,i} \}_{i=1}^n \leftarrow resample(\{ \mathbf{s}^r_{t,i}, \gamma_{t,i} \}_{i=1}^n)$ \\
      }
    } \label{alg:end_pf_om}
    \tcp{Return Mean Needle Pose}
    $\mathbf{s}^r_{t} \leftarrow \sum_{i=1}^{n} \gamma_{t,i} \mathbf{s}^r_{t,i}$ \\
    \label{alg:weight_sum}
  }
  \caption{Constrained Needle Tracking with HF/PF}
  \label{alg:constrained_needle_tracking}
\end{algorithm}

% In this section, we describe the detailed algorithms for integrating the feasibility constraints into a Bayesian filter with our proposed $(\alpha, w, u, v)$-state. 
Bayesian filtering is applied to solve (\ref{equ:cBF_formulation}) with the $(\alpha, w, u, v)$-state representation, which is easy to sample and constrain on $\mathcal{G}$.
Due to the nonlinearity in the observation models, Histogram Filter (HF) and Particle Filter (PF) are both implemented.
A summary is provided in Algorithm \ref{alg:constrained_needle_tracking}.

% For the HF, the feasibility constraints need to be imposed on a pre-collected discrete state space to ensure that the tracked states are always feasible. 
\textit{Initialization:} Upon initialization of the filters, a set of discrete states $\mathcal{S}^r$ such that $\mathcal{S}^r = \{ \mathbf{s}^r_1, \dots, \mathbf{s}^r_n | \mathbf{s}^r_i \in \mathcal{G}, \forall i \in \{1, \dots, n\} \}$ is collected.
We sample directly in our proposed $(\alpha, w, u, v)$-state space 
\begin{equation}
\mathbf{s}^r = [\alpha\ \ w\ \ u\ \ v]^\top, \text{~~~where~~~~} \label{equ:s_r_definition}
\end{equation}
\begin{align}
    & \alpha \sim \mathcal{U}\left( \frac{\pi}{2}, \frac{3\pi}{2} \right) \\
    & w \sim \mathcal{U}\left( d_{min}^3, d_{max}^3 \right) \\
    & u \sim \mathcal{U}\left( \frac{\theta_{min}}{2\pi}, \frac{\theta_{max}}{2\pi} \right) \\
    & v \sim \mathcal{U}\left( \frac{1}{2}(\cos\phi_{max}+1), \frac{1}{2}(\cos\phi_{min}+1) \right).
    \label{equ:sample_s_r}
\end{align}
$\mathcal{U}(\cdot, \cdot)$ is the uniform distribution.
The initialization is done in lines \ref{alg:initialize_sample} and \ref{alg:initialize_weights} of Algorithm \ref{alg:constrained_needle_tracking}.
% The same sampling approach is used to initialize the particles in the PF.
% 
% For the PF, the feasibility constraints need to be imposed on the motion model, $f(\cdot)$, such that its output always lies in $\mathcal{F}$. 
% To satisfy this condition, we apply (\ref{equ:s_r_motion_model}) with $\mathbf{w}^r_t \sim \Sigma^r_m$, where $\Sigma^r_m \in \mathbb{R}^{4 \times 4}$ is the covariance of the motion noise.

\textit{Motion Model:} The motion model is with zero-mean Gaussian noise since a grasped suture needle is not expected to move largely inside a gripper.
Therefore, the motion model's probability distribution is
\begin{equation}
    p_f(\cdot | \mathbf{s}^r_t, \mathbf{a}^r_t) \sim \mathcal{N}(\mathbf{s}^r_t + \mathbf{a}^r_t, \Sigma^r_m).
\end{equation}
where $\mathbf{a}^r_t = \mathbf{0}$ $\forall$ $t\in \{1, \dots, T\}$, and $\Sigma^r_m \in \mathbb{R}^{4 \times 4}$ is the covariance matrix of the motion noise.
The motion model is applied in lines \ref{alg:start_mm}-\ref{alg:end_mm} in Algorithm \ref{alg:constrained_needle_tracking}.

\textit{Observation Model:} The observation model is our previously proposed \textit{Points Matching to Ellipse Observation Model}~\cite{chiu2021markerless} which can be applied to any pixel detections of a suture needle.
Note that the observation model requires the pose-state, $\mathbf{s}^p_t = [\mathbf{b}^{c\top}_n\ \mathbf{q}^{c\top}_n]^\top$, as the input since the detections are from an image.
For more details on the observation model, see~\cite{chiu2021markerless}.
In Algorithm 1, the observation model is applied in lines \ref{alg:start_om}-\ref{alg:end_om}, where $\sigma_o^2 \in \mathbb{R}$ is the variance of the observation model.
Additional steps for PF, such as resampling particles, are detailed in Lines \ref{alg:start_pf_om}-\ref{alg:end_pf_om}.

% For a Bayesian filter that maintains a set of state candidates and their weights (probabilities), e.g., HF and PF, those candidates are usually averaged by their weights to obtain a more accurate tracking result.
% Therefore, the final tracked state at $t$, $\mathbf{s}^r_t$, becomes 
% \begin{equation}
%     \mathbf{s}^r_t = \sum_{\mathbf{x} \in \mathcal{S}^r} p_{t|t}(\mathbf{x}) \mathbf{x}
% \end{equation}
% for the HF and 
% \begin{equation}
%     \mathbf{s}^r_t = \sum_{i=1}^n \gamma_{t,i} \mathbf{s}^r_{t,i}
% \end{equation}
% for the PF.
\textit{Output:} To get a mean result from the HF and PF  algorithms, the weighted average of the discrete states is computed as shown in line \ref{alg:weight_sum} in Algorithm \ref{alg:constrained_needle_tracking}.
Note that $\mathbf{s}^r_t$ is still feasible since $\mathcal{G}$ is a convex space.
% If the states was represented by their 6D pose, then the final tracked state at $t$, $\mathbf{s}^p_t$, has to be chosen by the maximal weight (probability), i.e., $\mathbf{s}^p_t = \textrm{arg}\underset{\mathbf{x} \in \mathcal{S}^p}{\textrm{min}} p_{t|t}(\mathbf{x})$ for the HF and $\mathbf{s}^p_t = \mathbf{s}^p_{t,i}$, where $i = \textrm{arg}\underset{i \in \{1, \dots, n\}}{\textrm{min}} \gamma_{t,i}$ for the PF.

% We summarize the algorithms for tracking the 6D pose of an in-hand suture needle in Algorithm \ref{alg:constrained_needle_tracking}.

\section{Experiment and Results}

\begin{figure*}[t!]
    \vspace{1.5mm}
    \centering
    \begin{subfigure}{0.32\linewidth}
        \centering
        \includegraphics[width=\textwidth]{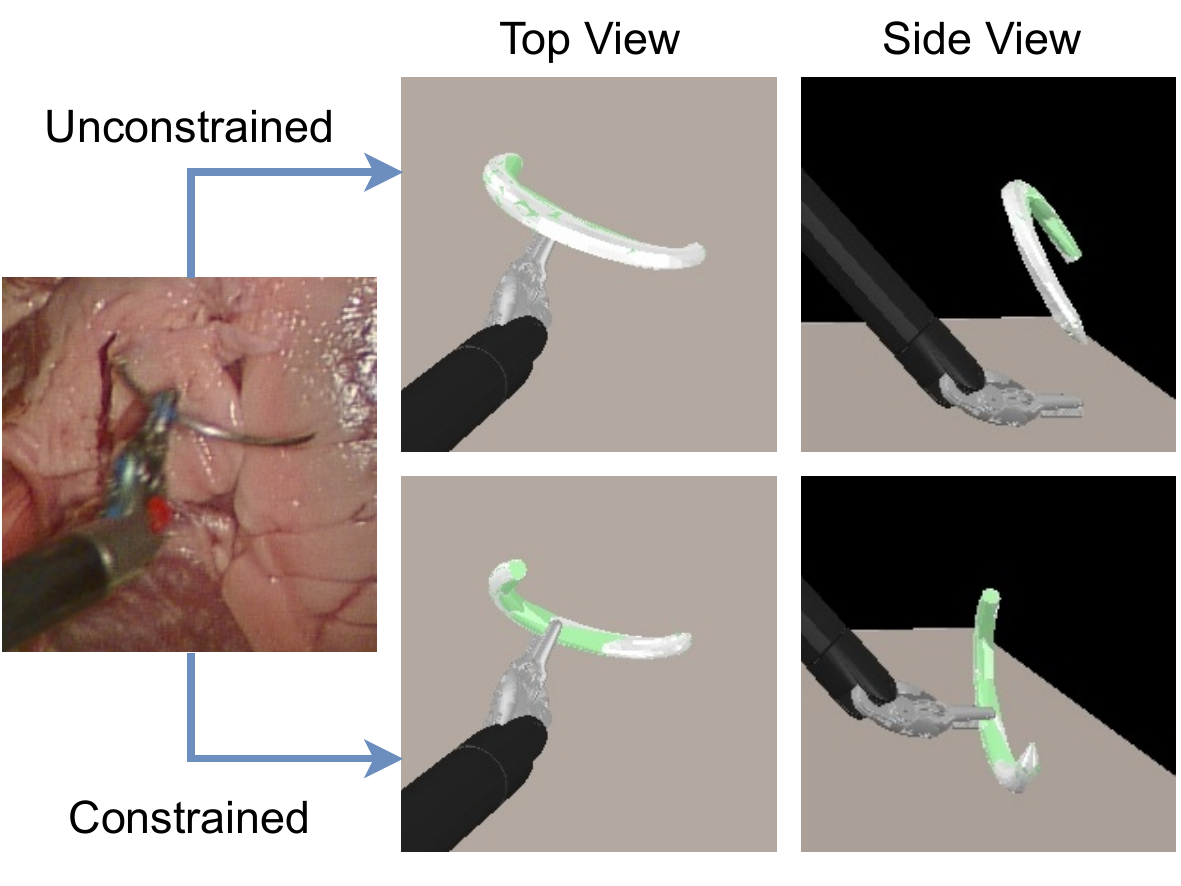}
    \end{subfigure}
    \begin{subfigure}{0.32\linewidth}
        \centering
        \includegraphics[width=\textwidth]{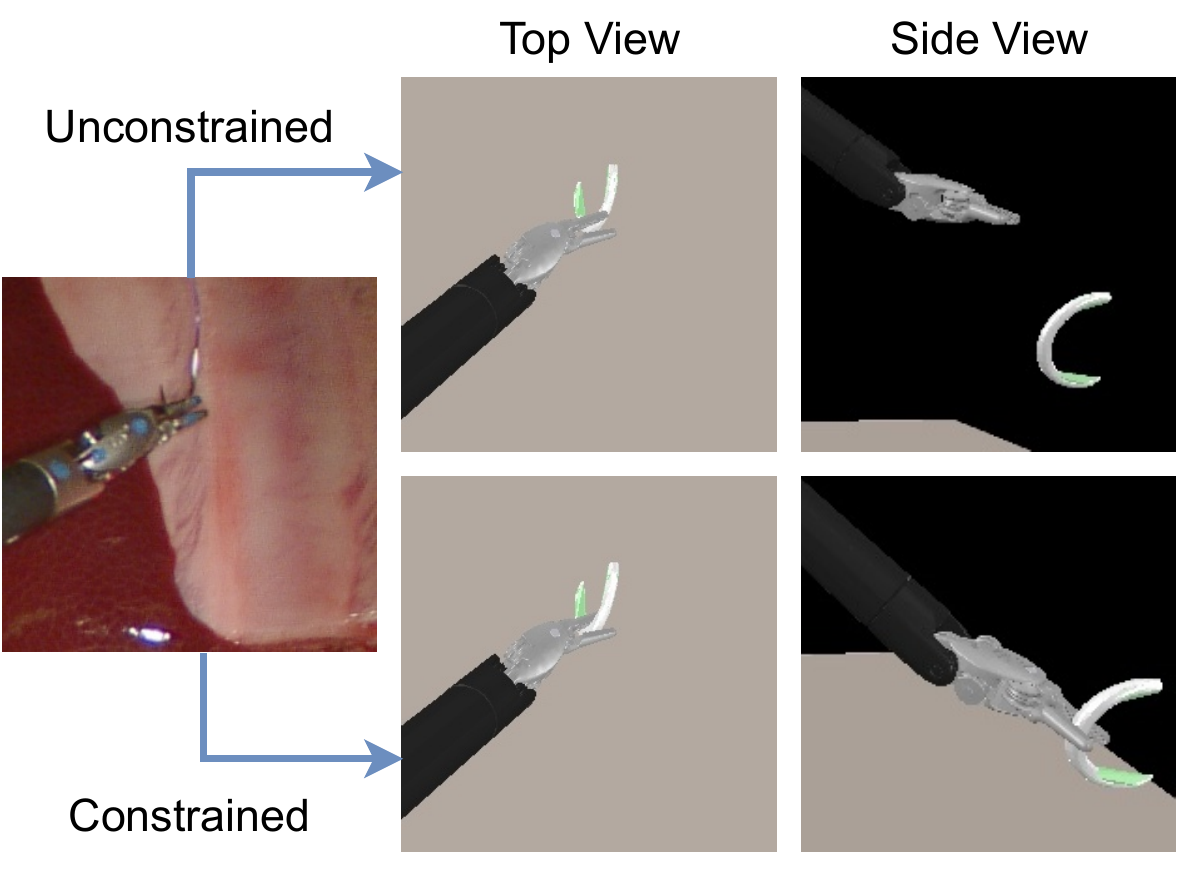}
    \end{subfigure}
    \begin{subfigure}{0.32\linewidth}
        \centering
        \includegraphics[width=\textwidth]{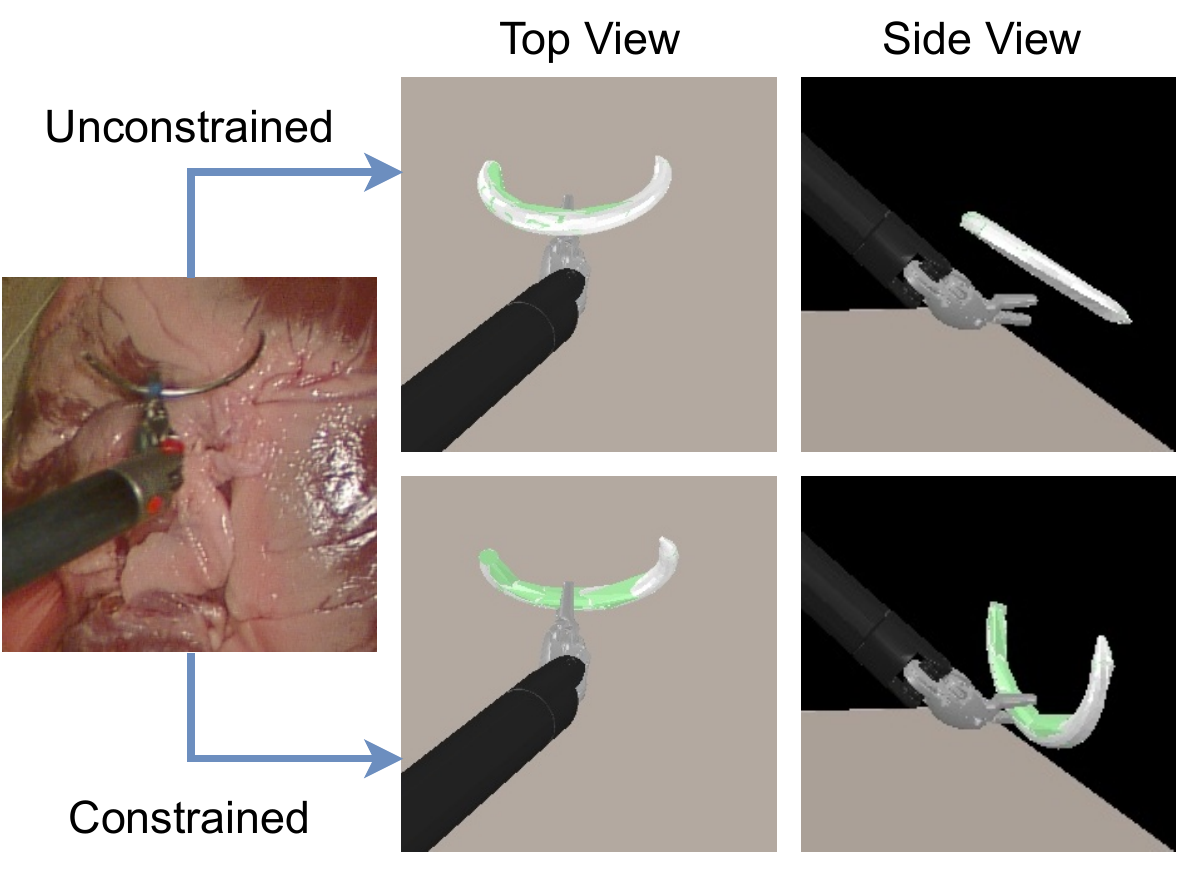}
    \end{subfigure}
    \caption{Raw image, top and side views of tool reconstruction from an unconstrained (PF) and our constrained (cPFrp) needle tracking approaches, across three examples. 
    Without incorporating grasping constraints into tracking, PF often estimates the needle pose with inaccurate depth and orientation, whereas cPFrp provides a more realistic, feasible reconstructed pose.} %more accurate in tracking the relative pose between the needle and the end-effector.}
    \label{fig:real_world_results}
    \vspace{-3.5mm}
\end{figure*}

\subsection{Tracking Experiments in Simulation}

We first evaluate our proposed constrained in-hand suture needle pose tracking algorithms in a CoppeliaSim environment.  
The simulation environmental settings are similar to the ones in~\cite{chiu2021markerless}, where a surgical manipulator holds a radius of 5.4mm suture needle. 
A stereo camera is set up to capture images with a size of $256 \times 256$, and five needle points are extracted from each image as the detections. 
Each detected point is disturbed by a noise sampled from $\mathcal{N}(\mathbf{0}, \Sigma_n)$, where $\Sigma_n \in \mathbb{R}^{2\times 2}$ is a diagonal matrix with each diagonal element being $\sigma_n^2$. 
Each experiment is repeated for 20 trials, and each trial contains 100 time steps.

We compare the following state definitions for an in-hand suture needle and tracking algorithms: 
\begin{enumerate}[leftmargin=*,align=left]
    \item Unconstrained pose-state, PF (PF)~\cite{chiu2021markerless,ozguner2018three}: naive PF with pose-states, and the needle motions are set to be the same as that of the end-effector. 
    \item Constrained pose-state by rejection sampling, HF (cHFrj)~\cite{boyko2021histogram}: constrained HF with feasible states pre-collected by rejection sampling. 
    \item Constrained pose-state by rejection sampling, PF (cPFrj)~\cite{lang2007bayesian}: constrained PF with rejection sampling applied to the outputs of the motion model. 
    \item Reparameterized $(\alpha, w, u, v)$-state, HF (cHFrp): our proposed state space with HF (Algorithm \ref{alg:constrained_needle_tracking}). 
    \item Reparameterized $(\alpha, w, u, v)$-state, PF (cPFrp): our proposed state space with PF (Algorithm \ref{alg:constrained_needle_tracking}).
\end{enumerate}
For all algorithms, we use the state-of-the-art \textit{Points Matching to Ellipse Observation Model} for needle detections~\cite{chiu2021markerless}. 
Each algorithm is run with 2000 particles or pre-collected states. 
We use the following two equations to calculate the positional and orientational errors of a tracked needle pose: 
\begin{equation}
    err_{\mathbf{b}} = \lVert \mathbf{b}_t - \overline{\mathbf{b}}_t \rVert_2,\ 
    err_{\mathbf{q}} = \lVert \mathbf{q}_t (\overline{\mathbf{q}}_t)^{-1} \rVert_2, 
\end{equation}
where $\overline{\cdot}$ is the ground truth obtained in the simulator.

Fig. \ref{fig:simulation_tracking_errors} shows the pose errors of tracking an in-hand suture needle while moving the end-effector. 
Although Bayesian filters with rejection sampling consider the feasible grasping constraints, their tracking errors are larger than other methods, even than the unconstrained one. 
This is because the feasible grasping manifold for pose-states is too irregular and difficult to sample uniformly from. 
With such an irregular space, the states sampled from it usually lack diversity since some states are easier to be sampled than others. 
Moreover, it requires lots of samples and feasibility checks to obtain enough feasible states. 
The average run time of cPFrj, which runs the rejection sampling process online, is 3.6 seconds per image frame, whereas other methods take 0.15 seconds per image frame. 

Tracking an in-hand needle with our reparameterized $(\alpha, w, u, v)$-states achieves the lowest pose errors. 
Sampling from the feasible grasping manifold of $(\alpha, w, u, v)$-states is much easier and requires no feasibility checks. 
Hence, both tracking errors and run time can be much lower than rejection-sampling-based methods. 
In addition, our proposed methods consider the physical interactions between the gripper and the needle. 
Therefore, when the environmental noise increases, the tracking errors of our methods remain low compared to the unconstrained method. 
Finally, the performance of cPFrp is better than cHFrp since PF allows online adjustment of the tracked state candidates. 
This adjustment moves those candidates closer to the real state. 

\subsection{Automated Suture Needle Regrasping}

\begin{table}[t]
\centering
\caption{Success rate for needle regrasping policy~\cite{chiu2021bimanual}.} 
%Our constrained method makes the policy more robust to noise in detections, indicating that it is crucial to consider the grasping constraints when automating suture needle manipulation tasks.}
\label{tab:regrasping_results}
\begin{tabular}{c|ccccc}
    \hline
    \multicolumn{6}{c}{$\sigma_{e,p} = 0$ mm, $\sigma_{e,o} = 0$ degree} \\
    \hline
    $\sigma_n$ (pixels) & 1 & 2 & 3 & 4 & 5 \\
    \hline
    PF & 0.9 & 0.77 & 0.6 & 0.6 & 0.43 \\
    cPFrp & 0.97 & 0.87 & 0.97 & 0.87 & 0.77 \\
    \hline\hline
    \multicolumn{6}{c}{$\sigma_{e,p} = 1$ mm, $\sigma_{e,o} = 5$ degrees} \\
    \hline
    $\sigma_n$ (pixels) & 1 & 2 & 3 & 4 & 5 \\
    \hline
    PF & 0.67 & 0.6 & 0.6 & 0.37 & 0.4 \\
    cPFrp & 0.83 & 0.9 & 0.9 & 0.87 & 0.87 \\
    \hline\hline
    \multicolumn{6}{c}{$\sigma_{e,p} = 2$ mm, $\sigma_{e,o} = 10$ degrees} \\
    \hline
    $\sigma_n$ (pixels) & 1 & 2 & 3 & 4 & 5 \\
    \hline
    PF & 0 & 0.07 & 0.03 & 0 & 0 \\
    cPFrp & 0.63 & 0.67 & 0.6 & 0.57 & 0.73 \\
    \hline
    \end{tabular}
\vspace{-5mm}
\end{table}

To demonstrate the importance of integrating grasping constraints into suture needle tracking, we compare the unconstrained PF and our proposed cPFrp methods on the suture needle regrasping task~\cite{chiu2021bimanual} in the simulation environment. 
In these experiments, we add noise not only to needle detections but also to the detected end-effector poses. 
The positional noise of end-effector poses is sampled from $\mathcal{N}(\mathbf{0}, \Sigma_{e,p})$, where $\Sigma_{e,p} \in \mathbb{R}^{3 \times 3}$ is a diagonal matrix with each diagonal element being $\sigma^2_{e,p}$. 
The orientational noise of the end-effector pose is $\theta_e [0\ 1\ 0]^\top$, where $\theta_e \in \mathbb{R}$ is sampled from $\mathcal{N}(0, \sigma^2_{e,o})$. 
Each experiment is run for 30 trials.

Table \ref{tab:regrasping_results} shows the regrasping success rate of PF and cPFrp under different environmental noise. 
PF barely shows any success when more noise is added to detected end-effector poses. 
On the other hand, cPFrp succeeds in multiple regrasping trials under the same condition. 
These results suggest that when automating suture needle manipulation tasks, it is essential to consider the relationship between the needle and the gripper that is interacting with it.

\subsection{Tracking Experiments in Real World}

We evaluate PF and our cPFrp methods on ex-vivo datasets and demonstrate the tool reconstruction results. 
The suture needles tested are with radii 7mm and 11.5mm, and the algorithms are run with 2000 particles. 
In the ex-vivo datasets, a needle is grasped by a Large Needle Driver (LND) installed on a Patient Side Manipulator (PSM) arm of the da Vinci Research Kit (dVRK)~\cite{kazanzides2014open}. 
dVRK’s stereo endoscopic camera, which is 1080p and runs with 30 fps, is used to capture images for end-effector and needle detections. 
The end-effector poses are tracked by our previous method~\cite{richter2021robotic}, and the markerless needle detections are obtained by DeepLabCut~\cite{mathis2018deeplabcut}, the state-of-the-art keypoint detector.

Figure \ref{fig:cover_image} and \ref{fig:real_world_results} show the top and side views of tool reconstruction. 
From the top views, the results of both the unconstrained and constrained methods align well with the raw images. 
However, the side views clearly show their differences. 
The unconstrained method can lead to the reconstructed needle floating in the air or in collision with the gripper. 
Also, the inaccuracy mainly happens when reconstructing the depths of a needle from the camera. 
This is because the detections of a needle with different depths do not show many differences in the endoscopic images. 
Therefore, without integrating the feasibility constraints into needle tracking, the unconstrained method is likely to estimate a needle pose with inaccurate depth.

Our proposed constrained method ensures the needle pose is always feasible given the estimated end-effector pose. 
Note that although there can be noise in end-effector poses, the reconstructed relative pose between the needle and the end-effector is accurate for our constrained method. 
In Table \ref{tab:regrasping_results}, we show that the accuracy of this relative pose is more crucial in automating suture needle manipulation tasks.

\section{Discussion and Conclusion}

In this work, we propose approaches to incorporate feasible grasping constraints into real-time localizing the 6D pose of an in-hand suture needle.
We define a novel state space for the 6D pose of a needle. 
This state space allows efficient sampling from the feasible grasping manifold, which requires no feasibility checks with external software. 
We incorporate the proposed state space and feasible grasping constraints into Bayesian filters for real-time needle pose tracking and demonstrate that our constrained approaches outperform previous ones. 
Our methods focus on tracking the relative poses between the needle and the gripper holding it and rely more on the end-effector poses. 
However, a surgical manipulator can be tracked accurately by previous methods~\cite{richter2021robotic}, and we show in our experiments that accurately tracking the relative poses is more crucial in successfully automating suture needle manipulation tasks.

%\input{sections/draft}

%\clearpage
\balance
\bibliographystyle{IEEEtran}
\bibliography{ref}

\begin{thebibliography}{10}
\providecommand{\url}[1]{#1}
\csname url@samestyle\endcsname
\providecommand{\newblock}{\relax}
\providecommand{\bibinfo}[2]{#2}
\providecommand{\BIBentrySTDinterwordspacing}{\spaceskip=0pt\relax}
\providecommand{\BIBentryALTinterwordstretchfactor}{4}
\providecommand{\BIBentryALTinterwordspacing}{\spaceskip=\fontdimen2\font plus
\BIBentryALTinterwordstretchfactor\fontdimen3\font minus
  \fontdimen4\font\relax}
\providecommand{\BIBforeignlanguage}[2]{{%
\expandafter\ifx\csname l@#1\endcsname\relax
\typeout{** WARNING: IEEEtran.bst: No hyphenation pattern has been}%
\typeout{** loaded for the language `#1'. Using the pattern for}%
\typeout{** the default language instead.}%
\else
\language=\csname l@#1\endcsname
\fi
#2}}
\providecommand{\BIBdecl}{\relax}
\BIBdecl

\bibitem{yip2019robot}
M.~Yip and N.~Das, ``Robot autonomy for surgery,'' in \emph{The Encyclopedia of
  MEDICAL ROBOTICS: Volume 1 Minimally Invasive Surgical Robotics}.\hskip 1em
  plus 0.5em minus 0.4em\relax World Scientific, 2019, pp. 281--313.

\bibitem{garcia1998manual}
A.~Garcia-Ruiz, M.~Gagner, J.~H. Miller, C.~P. Steiner, and J.~F. Hahn,
  ``Manual vs robotically assisted laparoscopic surgery in the performance of
  basic manipulation and suturing tasks,'' \emph{Archives of surgery}, vol.
  133, no.~9, pp. 957--961, 1998.

\bibitem{hubens2003performance}
G.~Hubens, H.~Coveliers, L.~Balliu, M.~Ruppert, and W.~Vaneerdeweg, ``A
  performance study comparing manual and robotically assisted laparoscopic
  surgery using the da vinci system,'' \emph{Surgical Endoscopy and other
  interventional techniques}, vol.~17, no.~10, pp. 1595--1599, 2003.

\bibitem{corcione2005advantages}
F.~Corcione, C.~Esposito, D.~Cuccurullo, A.~Settembre, N.~Miranda, F.~Amato,
  F.~Pirozzi, and P.~Caiazzo, ``Advantages and limits of robot-assisted
  laparoscopic surgery: preliminary experience,'' \emph{Surgical Endoscopy and
  Other Interventional Techniques}, vol.~19, no.~1, pp. 117--119, 2005.

\bibitem{li2020super}
Y.~Li, F.~Richter, J.~Lu, E.~K. Funk, R.~K. Orosco, J.~Zhu, and M.~C. Yip,
  ``Super: A surgical perception framework for endoscopic tissue manipulation
  with surgical robotics,'' \emph{IEEE Robotics and Automation Letters},
  vol.~5, no.~2, pp. 2294--2301, 2020.

\bibitem{lu2021super}
J.~Lu, A.~Jayakumari, F.~Richter, Y.~Li, and M.~C. Yip, ``Super deep: A
  surgical perception framework for robotic tissue manipulation using deep
  learning for feature extraction,'' in \emph{2021 IEEE International
  Conference on Robotics and Automation (ICRA)}.\hskip 1em plus 0.5em minus
  0.4em\relax IEEE, 2021, pp. 4783--4789.

\bibitem{richter2021robotic}
F.~Richter, J.~Lu, R.~K. Orosco, and M.~C. Yip, ``Robotic tool tracking under
  partially visible kinematic chain: A unified approach,'' \emph{IEEE
  Transactions on Robotics}, 2021.

\bibitem{chiu2021bimanual}
Z.-Y. Chiu, F.~Richter, E.~K. Funk, R.~K. Orosco, and M.~C. Yip, ``Bimanual
  regrasping for suture needles using reinforcement learning for rapid motion
  planning,'' in \emph{2021 IEEE International Conference on Robotics and
  Automation (ICRA)}.\hskip 1em plus 0.5em minus 0.4em\relax IEEE, 2021, pp.
  7737--7743.

\bibitem{wilcox2022learning}
A.~Wilcox, J.~Kerr, B.~Thananjeyan, J.~Ichnowski, M.~Hwang, S.~Paradis, D.~Fer,
  and K.~Goldberg, ``Learning to localize, grasp, and hand over unmodified
  surgical needles,'' in \emph{2022 International Conference on Robotics and
  Automation (ICRA)}.\hskip 1em plus 0.5em minus 0.4em\relax IEEE, 2022, pp.
  9637--9643.

\bibitem{chow2013improved}
D.-L. Chow and W.~Newman, ``Improved knot-tying methods for autonomous robot
  surgery,'' in \emph{2013 IEEE International Conference on Automation Science
  and Engineering (CASE)}.\hskip 1em plus 0.5em minus 0.4em\relax IEEE, 2013,
  pp. 461--465.

\bibitem{lu2018vision}
B.~Lu, H.~K. Chu, K.~Huang, and L.~Cheng, ``Vision-based surgical suture
  looping through trajectory planning for wound suturing,'' \emph{IEEE
  Transactions on Automation Science and Engineering}, vol.~16, no.~2, pp.
  542--556, 2018.

\bibitem{richter2021autonomous}
F.~Richter, S.~Shen, F.~Liu, J.~Huang, E.~K. Funk, R.~K. Orosco, and M.~C. Yip,
  ``Autonomous robotic suction to clear the surgical field for hemostasis using
  image-based blood flow detection,'' \emph{IEEE Robotics and Automation
  Letters}, vol.~6, no.~2, pp. 1383--1390, 2021.

\bibitem{iyer2013single}
S.~Iyer, T.~Looi, and J.~Drake, ``A single arm, single camera system for
  automated suturing,'' in \emph{2013 IEEE International Conference on Robotics
  and Automation}.\hskip 1em plus 0.5em minus 0.4em\relax IEEE, 2013, pp.
  239--244.

\bibitem{ferro2017vision}
M.~Ferro, G.~Fontanelli, F.~Ficuciello, B.~Siciliano, and M.~Vendittelli,
  ``Vision-based suturing needle tracking with extended kalman filter,'' in
  \emph{Computer/Robot Assisted Surgery workshop}, 2017.

\bibitem{sen2016automating}
S.~Sen, A.~Garg, D.~V. Gealy, S.~McKinley, Y.~Jen, and K.~Goldberg,
  ``Automating multi-throw multilateral surgical suturing with a mechanical
  needle guide and sequential convex optimization,'' in \emph{2016 IEEE
  international conference on robotics and automation (ICRA)}.\hskip 1em plus
  0.5em minus 0.4em\relax IEEE, 2016, pp. 4178--4185.

\bibitem{d2018automated}
C.~D'Ettorre, G.~Dwyer, X.~Du, F.~Chadebecq, F.~Vasconcelos, E.~De~Momi, and
  D.~Stoyanov, ``Automated pick-up of suturing needles for robotic surgical
  assistance,'' in \emph{2018 IEEE International Conference on Robotics and
  Automation (ICRA)}.\hskip 1em plus 0.5em minus 0.4em\relax IEEE, 2018, pp.
  1370--1377.

\bibitem{kurose2013preliminary}
Y.~Kurose, Y.~M. Baek, Y.~Kamei, S.~Tanaka, K.~Harada, S.~Sora, A.~Morita,
  N.~Sugita, and M.~Mitsuishi, ``Preliminary study of needle tracking in a
  microsurgical robotic system for automated operations,'' in \emph{2013 13th
  international conference on control, automation and systems (ICCAS
  2013)}.\hskip 1em plus 0.5em minus 0.4em\relax IEEE, 2013, pp. 627--630.

\bibitem{ozguner2018three}
O.~{\"O}zg{\"u}ner, R.~Hao, R.~C. Jackson, T.~Shkurti, W.~Newman, and M.~C.
  Cavusoglu, ``Three-dimensional surgical needle localization and tracking
  using stereo endoscopic image streams,'' in \emph{2018 IEEE international
  conference on robotics and automation (ICRA)}.\hskip 1em plus 0.5em minus
  0.4em\relax IEEE, 2018, pp. 6617--6624.

\bibitem{chiu2021markerless}
Z.-Y. Chiu, A.~Z. Liao, F.~Richter, B.~Johnson, and M.~C. Yip, ``Markerless
  suture needle 6d pose tracking with robust uncertainty estimation for
  autonomous minimally invasive robotic surgery,'' \emph{arXiv preprint
  arXiv:2109.12722}, 2021.

\bibitem{lang2007bayesian}
L.~Lang, W.-s. Chen, B.~R. Bakshi, P.~K. Goel, and S.~Ungarala, ``Bayesian
  estimation via sequential monte carlo sampling—constrained dynamic
  systems,'' \emph{Automatica}, vol.~43, no.~9, pp. 1615--1622, 2007.

\bibitem{kong1994sequential}
A.~Kong, J.~S. Liu, and W.~H. Wong, ``Sequential imputations and bayesian
  missing data problems,'' \emph{Journal of the American statistical
  association}, vol.~89, no. 425, pp. 278--288, 1994.

\bibitem{zhao2014constrained}
Z.~Zhao, B.~Huang, and F.~Liu, ``Constrained particle filtering methods for
  state estimation of nonlinear process,'' \emph{AIChE Journal}, vol.~60,
  no.~6, pp. 2072--2082, 2014.

\bibitem{hu2020particle}
C.~Hu, Y.~Liang, X.~Wang, and L.~Xu, ``A particle filter via constrained
  sampling for nonlinear dynamic systems,'' \emph{International Journal of
  Robust and Nonlinear Control}, vol.~30, no.~13, pp. 4944--4959, 2020.

\bibitem{hasson2019learning}
Y.~Hasson, G.~Varol, D.~Tzionas, I.~Kalevatykh, M.~J. Black, I.~Laptev, and
  C.~Schmid, ``Learning joint reconstruction of hands and manipulated
  objects,'' in \emph{Proceedings of the IEEE/CVF conference on computer vision
  and pattern recognition}, 2019, pp. 11\,807--11\,816.

\bibitem{weisstein2002sphere}
E.~W. Weisstein, ``Sphere point picking,'' \emph{https://mathworld. wolfram.
  com/}, 2002.

\bibitem{cundy1989sphere}
H.~Cundy and A.~Rollett, ``Sphere and cylinder----archimedes’ theorem,''
  \emph{Mathematical Models}, pp. 172--173, 1989.

\bibitem{harman2010decompositional}
R.~Harman and V.~Lacko, ``On decompositional algorithms for uniform sampling
  from n-spheres and n-balls,'' \emph{Journal of Multivariate Analysis}, vol.
  101, no.~10, pp. 2297--2304, 2010.

\bibitem{kitagawa1996monte}
G.~Kitagawa, ``Monte carlo filter and smoother for non-gaussian nonlinear state
  space models,'' \emph{Journal of computational and graphical statistics},
  vol.~5, no.~1, pp. 1--25, 1996.

\bibitem{boyko2021histogram}
N.~Boyko and Y.~Hladun, ``Histogram filter for robot localization,'' in
  \emph{2021 IEEE 16th International Conference on Computer Sciences and
  Information Technologies (CSIT)}, vol.~1.\hskip 1em plus 0.5em minus
  0.4em\relax IEEE, 2021, pp. 38--43.

\bibitem{kazanzides2014open}
P.~Kazanzides, Z.~Chen, A.~Deguet, G.~S. Fischer, R.~H. Taylor, and S.~P.
  DiMaio, ``An open-source research kit for the da vinci{\textregistered}
  surgical system,'' in \emph{2014 IEEE international conference on robotics
  and automation (ICRA)}.\hskip 1em plus 0.5em minus 0.4em\relax IEEE, 2014,
  pp. 6434--6439.

\bibitem{mathis2018deeplabcut}
A.~Mathis, P.~Mamidanna, K.~M. Cury, T.~Abe, V.~N. Murthy, M.~W. Mathis, and
  M.~Bethge, ``Deeplabcut: markerless pose estimation of user-defined body
  parts with deep learning,'' \emph{Nature neuroscience}, vol.~21, no.~9, pp.
  1281--1289, 2018.

\end{thebibliography}

\end{document}